\documentclass[10pt,twocolumn,letterpaper]{article}

\usepackage{iccv}
\usepackage{amsmath}
\usepackage{times}
\usepackage{epsfig}
\usepackage{graphicx}
\usepackage{amssymb}
\usepackage{multirow}

\usepackage[normalem]{ulem}

\usepackage[pagebackref=true,breaklinks=true,letterpaper=true,colorlinks,bookmarks=false]{hyperref}

\iccvfinalcopy 

\def\iccvPaperID{2195} 

\ificcvfinal\fi
\begin{document}

\title{Co-interest Person Detection from Multiple Wearable Camera Videos}

\author{Yuewei Lin$^1$ \quad Kareem Ezzeldeen$^1$ \quad Youjie Zhou$^1$ \quad Xiaochuan Fan$^1$ \\
Hongkai Yu$^1$ \quad Hui Qian$^2$ \quad Song Wang$^1$ \\
$^1$University of South Carolina, Columbia, SC 29208\\
$^2$Zhejiang University, Hangzhou, China\\
{\tt\small ywlin.cq@gmail.com}
}


\maketitle


\begin{abstract}
Wearable cameras, such as Google Glass and Go Pro, enable video data collection over larger areas and from different views. In this paper, we tackle a new problem of locating the co-interest person (CIP), i.e., the one who draws attention from most camera wearers, from temporally synchronized videos taken by multiple wearable cameras. Our basic idea is to exploit the motion patterns of people and use them to correlate the persons across different videos, instead of performing appearance-based matching as in traditional video co-segmentation/localization. This way, we can identify CIP even if a group of people with similar appearance are present in the view. More specifically, we detect a set of persons on each frame as the candidates of the CIP and then build a Conditional Random Field (CRF) model to select the one with consistent motion patterns in different videos and high spacial-temporal consistency in each video. We collect three sets of wearable-camera videos for testing the proposed algorithm. All the involved people have similar appearances in the collected videos and the experiments demonstrate the effectiveness of the proposed algorithm.
\end{abstract}

\section{Introduction}

Video-based individual, interactive, and group activity recognition has attracted more and more interests in the computer vision community. Using fixed cameras for collecting videos suffers from the problem of only covering very limited areas. This problem will get even worse when recognizing activities in a social event, such as a concert, ceremony or party, where multiple people are present and move from time to time. Recently, wearable cameras, such as Google Glass or Go Pro, provide a new solution, where all or part of the involved persons wear a camera over head to record what they see over time~\cite{Fathi12,Park14}.


By combining the temporally synchronized videos from different wearers, we can recognize the activity occurred in a large area, because camera wearers can walk or move the head to follow the people or event of interest~\cite{Zheng14}. An important problem arising from this setting is to identify the co-interest person (CIP) that attracts the attentions from multiple wearers since this person usually plays a central role in ongoing event of interest. The CIP and his/her activities are of particular importance for surveillance, anomaly detection and social network construction. For examples, in a public scenario such as an airport, CIP can be a person with abnormal behavior or activity who usually draws attention from multiple camera-wearing security guards and the quick detection of such CIPs can promote the public security. In a kindergarten, CIP may be a kid with strange behavior that continuously draws joint attentions from camera-wearing teachers or other kids. In this case, the CIP detection can facilitate the early findings of various child development issues. In a group discussion, people usually focus on the person who leads or gives the speech at any time and the identification of such CIPs over time can help summarize and edit all the videos from the attendee's cameras for more effective information management and retrieval. In this paper, we develop a new approach to detect CIPs from multiple videos taken by wearable cameras.

\begin{figure}[htbp]
\centering \includegraphics[width=1\columnwidth]{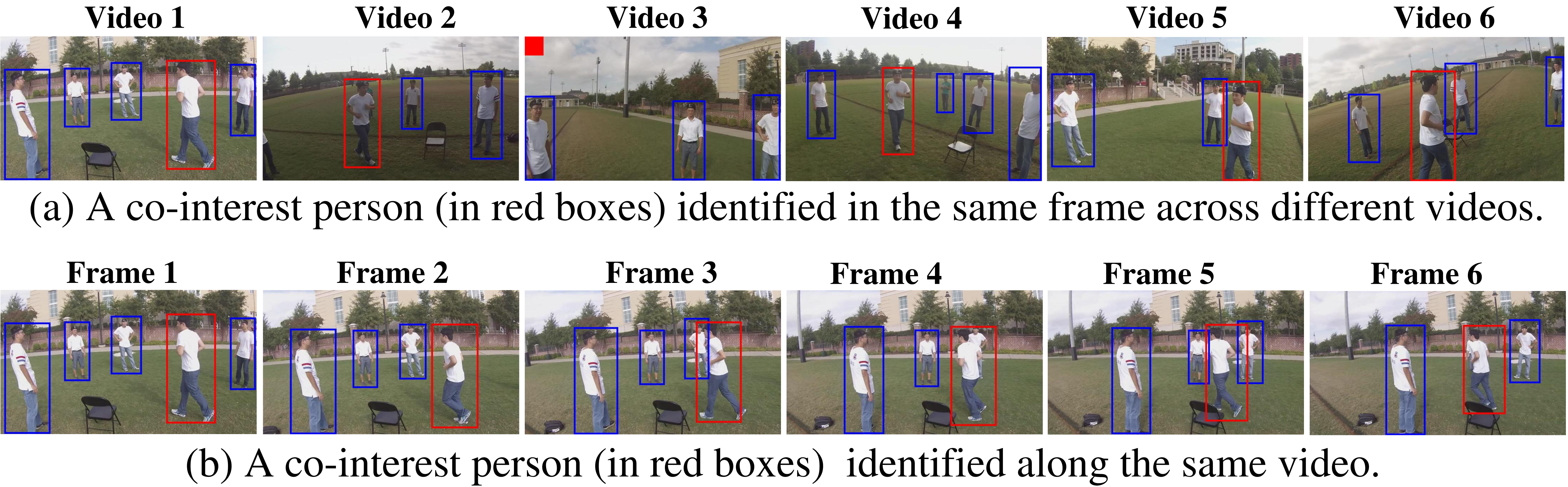} \caption{An illustration of the basic idea underlying the proposed CIP detection  approach. (a) A CIP (in red boxes) always shows consistent 3D motion patterns across all the videos in which he/she is present. (b) A CIP (in red boxes) usually shows high spatial-temporal consistency along a video. Our proposed algorithm considers the consistency  in both (a) and (b) for CIP detection. Note that the Video 3 in (a) is an egocentric video of the CIP.}
\label{fig:fig1}
\end{figure}


In many social events, attendees may wear clothes with similar color and texture, such as wearing specific uniforms in work and suits in a formal dinner. In these cases, it is very difficult to identify CIPs by performing appearance matching across multiple videos, as shown in Fig.~\ref{fig:fig1}(a). In this paper, we identify the CIP based on his/her motion patterns: it is unlikely that two persons in the view keep showing exactly same motion over time.

However, it is a very challenging problem to identify the person with the same motion pattern from different videos even if these videos are temporally synchronized, because the motion pattern of a person is defined in 3D and can only be partially reflected in each 2D video. In practice, the 3D motion of a same person may be projected to completely different 2D motions in different videos, as illustrated in Fig.1(a). In addition, in this research, the inference of the 2D motion pattern of a person is further complicated by the use of the wearable cameras: camera motion and person motion are mixed in generating each video.

In this paper, we address this challenging problem by combining the temporally synchronized frames from different videos using a Conditional Random Field (CRF) model. We first perform human detection to obtain a set of candidates of the CIP. Then we build a CRF by taking each frame as a node and the candidates on that frame as its states. In this CRF, we define an inter-video energy that reflects the motion-pattern difference of the candidates drawn from different videos, as illustrated in Fig.~\ref{fig:fig1}(a). In particular, we use histogram of optical flow (HoF), Hankelets~\cite{Li12} and motion pattern histograms (MPH)~\cite{Ciptadi14} to describe the human motion. We also include an intra-video energy term in the CRF to measure the location and size consistency of candidates across frames of a same video, as illustrated in Fig.~\ref{fig:fig1}(b). The minimization of the proposed CRF energy will generate a CIP on each frame of each video that shows both inter-video and intra-video properties. To handle the case where a frame contains no CIP, e.g., the CIP can not see himself in his egocentric video, as shown by video 3 in Fig.1(a), we also introduce an idle state in each frame.


\section{Related Work}

\subsection{Video co-segmentation}
Related to this paper is a series of prior research on video co-segmentation, where common objects are segmented from multiple videos. Video co-segmentation can be treated as an extension of the long-studied image co-segmentation~\cite{Dai13,Joulin10,Joulin12,Kim11,Li14,Meng13,Rubinstein13,Tang14,Vicente11,Wang14TIP}, where the input is a set of images instead of videos.

However, different from the proposed CIP detection, the multiple videos used for video co-segmentation are usually not temporarily synchronized: they may record the same object at different time. As a result, the co-segmented person may not show motion consistency across different videos. In practice, almost all the existing co-segmentation algorithms  are based on object-appearance matching. For example, \cite{Chen12} and \cite{Rubio12} model the co-segmentation as a foreground/background separation problem based on the appearance information. Wang et al. \cite{Wang14ECCV} develop an appearance based weakly supervised co-segmentation algorithm which also needs the labels for a few frames. In~\cite{Joulin14}, the common objects are localized in different videos by using the appearance and local features.

Some of prior video co-segmentation methods use the motion information to help track and/or segment the objects in each video but not corresponding objects across videos as in the proposed CIP detection. Chiu and Fritz \cite{Chiu13} propose a multi-class co-segmentation algorithm based on a non-parametric Bayesian model which uses the motion information for object segmentation. In \cite{Zhang14}, a number of tracklets are detected inside each video and the appearance and shape information along the tracklets are then extracted to identify the common target in multiple videos. In \cite{Fu14}, co-segmentation is formulated as a co-selection graph where motions are estimated to measure the spatial temporal consistency. In \cite{Guo13}, motion trajectories are detected to match the action across video pairs. However, the action matching is only in the high-level of the action type. There is no frame-by-frame motion consistency between these videos since they are not temporally synchronized.


In addition, when multiple people are present in  the view of each video, most works on video co-segmentation identify all of them as a common object -- person. In the proposed CIP detection, we need to distinguish them and identify one person with presence in all or most of the videos.

\subsection{Gaze concurrences}

Also related to our work is the research on gaze concurrences of multiple video takers. Robertson and Reid \cite{Robertson06} estimate face orientation by learning 2D face features from different views. In \cite{Smith08}, the points of interest are estimated in a crowded scene. However, these methods rely on video data captured from a third person. As a result, the area covered by these videos are quite limited and the accuracy of head pose estimation degrades when distance to the camera increases \cite{Park12}. Park et al. present an algorithm to locate gaze concurrences directly from videos taken by head-mounted cameras. However, this algorithm requires a prior scanning of the area of interest (for example, room or an auditorium) to reconstruct the reference structure. This may not be available in practice.


\section{Proposed Method}

\begin{figure}[htbp]
\centering \includegraphics[width=1\columnwidth]{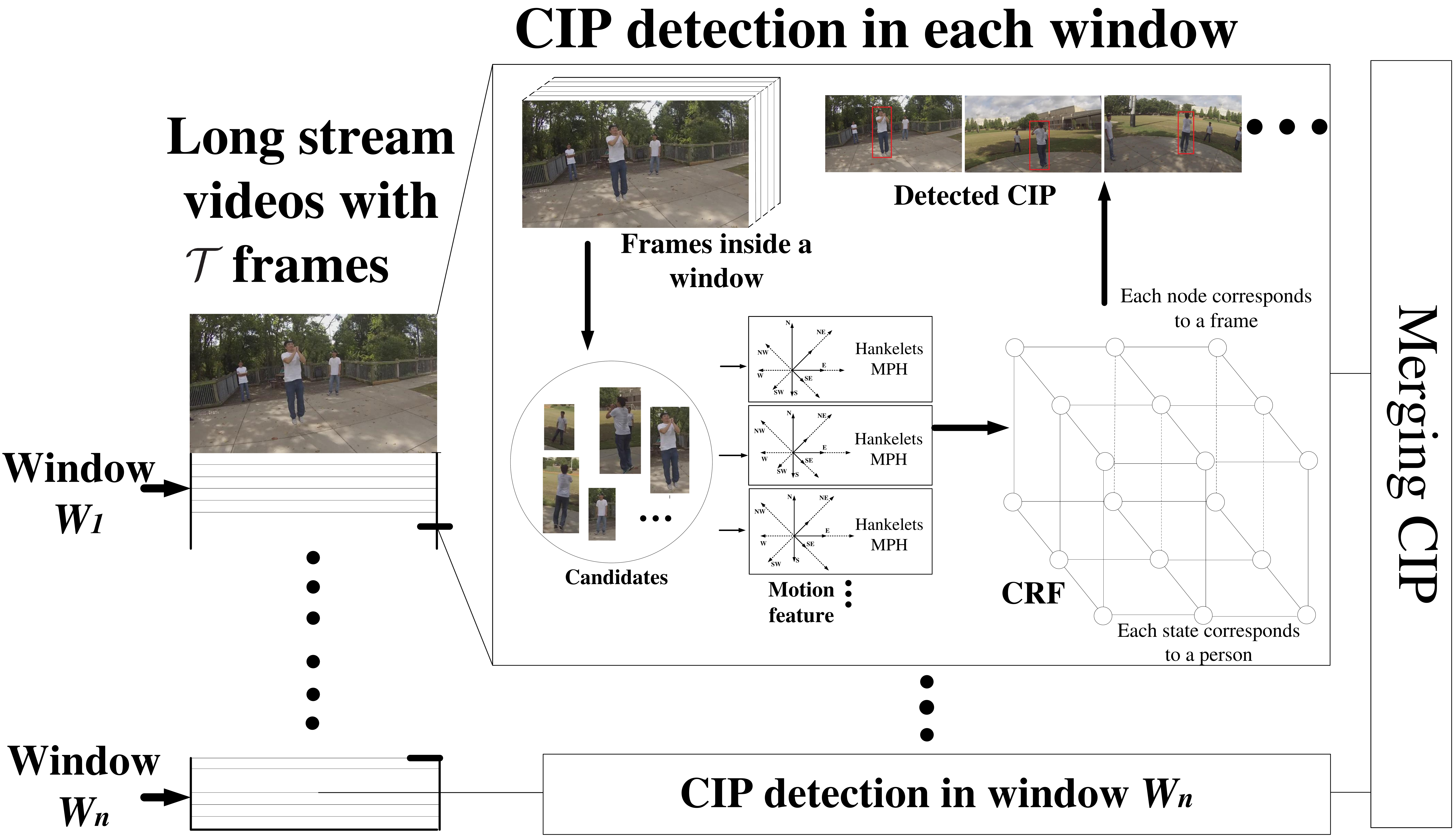} \caption{The framework of the proposed algorithm.}
\label{fig:Model}
\end{figure}

To detect CIP over time, we record a set of $N$ temporally synchronized long-streaming videos that are taken by $N$ wearable cameras over time $[0, \mathcal{T}]$. The CIP in these videos may change over time. To simplify the problem, we first apply a sliding window technique to divide the time $[0,\mathcal{T}]$ into overlapped short time windows with length $T$. Over each short time window, we assume that the CIP does not change in these $N$ videos and we propose an algorithm to detect such a person in each window. The proposed algorithm also provides an energy for the CIP detection in each window. This energy value negatively reflected the confidence of CIP detection. Finally, we merge the CIP detection results over all the windows based on their energies to achieve a CIP detection at each frame over $[0,\mathcal{T}]$, as illustrated in Fig.~\ref{fig:Model}.


To merge CIP detection results from all the windows, we always select the one with lowest detection energy at each frame. Specifically, by using sliding window technique, the constructed windows are partially overlapped and each frame, say $t$, is covered by multiple windows, say $W_1, W_2, \cdots, W_K$. In each window $W_k$, the CIP detection algorithm (to be introduced in Section~\ref{sec:CRFConstruction}) generates a CIP detection $P_k$ and an associated energy $E_k$. We find the one with the lowest energy as
\begin{equation}
k^*=\arg\min_{1\leq k\leq K} E_k
\end{equation}
and set $P_{k^*}$ as the final CIP detection in this frame $t$.

\begin{figure}[htbp]
\centering \includegraphics[width=1\columnwidth]{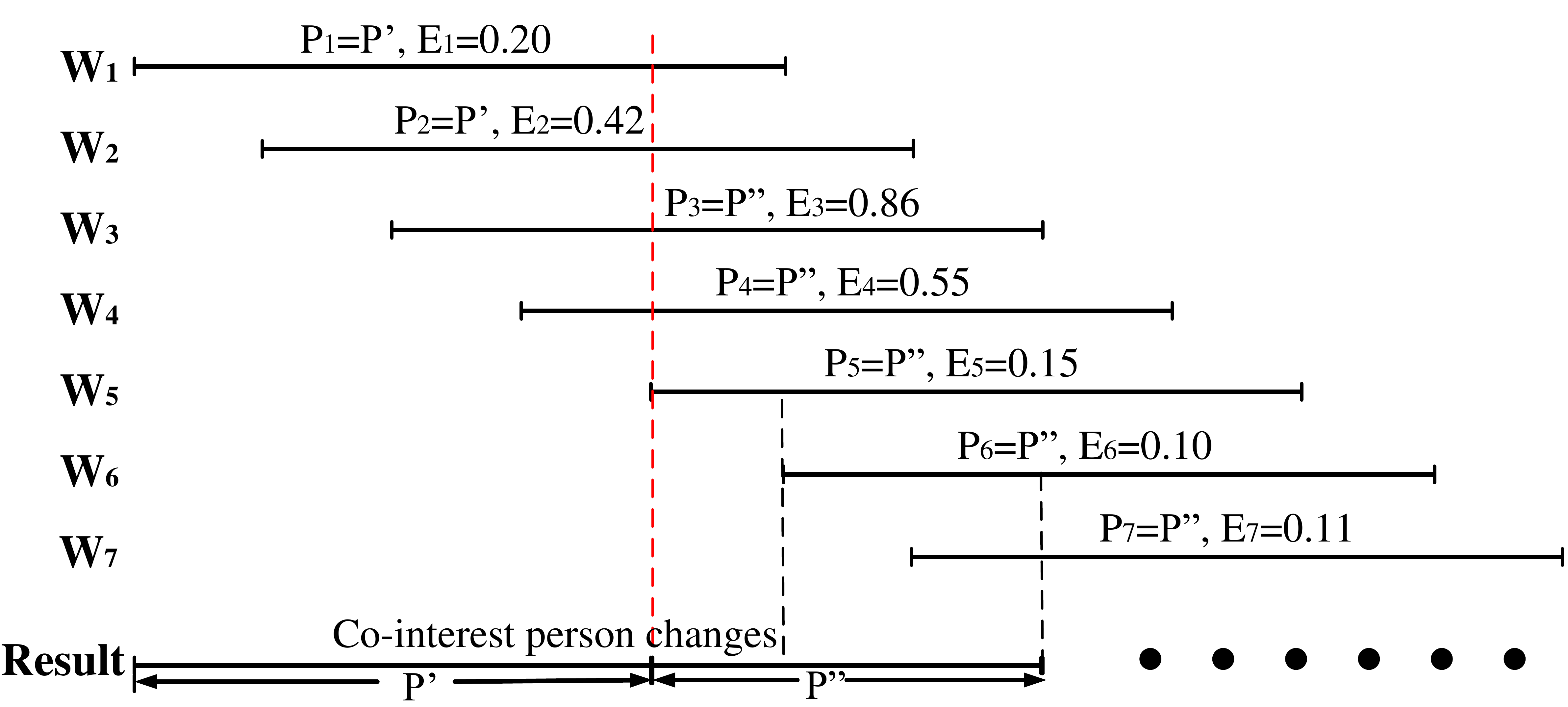}
\caption{An example that illustrates the merging of the CIP detection results.}
\label{fig:slidingWin}
\end{figure}

An example is shown in Fig.~\ref{fig:slidingWin}. In this figure, $W_i$ denotes the partially overlapped windows, and $P_i$ and $E_i$ denote the CIP detected in each window $W_i$ and its energy, respectively. If $P_1=P_2=P'$ and $P_3=P_4=P_5=P_6=P_7=P''$, as shown in Fig.~\ref{fig:slidingWin}, then the red dashed line actually indicates a time when CIP is changed from $P'$ to $P''$.





In the following, we focus on developing the proposed CIP detection algorithm in each window $W$.

\subsection{CIP detection using a CRF model}
\label{sec:CRFConstruction}

Over a short-time window $W$, the $N$ input videos are actually cropped into $N$ synchronized short video clips $\mathcal{F}=\{\mathcal{F}^n|n=1,2,\cdots, N\}$ with $\mathcal{F}^n=\{F_{t}^{n}|t=1,\cdots,T\}$ where $F_{t}^{n}$ is the $t$-th frame in the $n$-th video clip.

As shown in Fig.~\ref{fig:Model}, we first perform the human detection on each frame and take each detection as a CIP candidate. A conditional random field (CRF)~\cite{Deselaers12} is then constructed by treating each frame as a node and each candidate on this frame as a state of this node. Using this CRF model,
our goal is to seek a candidate $h_{t}^{n}$ on each frame $F_{t}^{n}$ as the detected CIP. Specifically, the CIP detection $H=\{h_{t}^{n}|n=1,\cdots,N;t=1,\cdots,T\}$
has a posterior probability
\begin{equation}
\begin{split}&p(H|\mathcal{F})\varpropto\exp(-E(H|\mathcal{F}))\\
\mathrm{with}\,\, &E(H|\mathcal{F})=\sum_{n,m,t,r}\Psi(h_{t}^{n},h_{r}^{m}|F_{t}^{n},F_{r}^{m}),
\end{split}
\label{eq:CRF}
\end{equation}
where $\Psi(h_{t}^{n},h_{r}^{m}|F_{t}^{n},F_{r}^{m})$ is a energy of matching $h_{t}^{n}$ and $h_{r}^{m}$ as the same person and taking it as the CIP.
In the remainder of the paper, we simplify the notation of this pairwise energy as $\Psi(h_{t}^{n},h_{r}^{m})$ and the energy function $E(H|\mathcal{F})$ as $E(H)$ when there is no ambiguity.
This way, the CIP detection in the short time window is reduced to a problem of finding an optimal $H$ that minimizes the energy $E(H|\mathcal{F})$.

The major problem to be solved here is the definition of the pairwise energy $\Psi(h_{t}^{n},h_{r}^{m})$, which should reflect the correspondence of the CIP between a pair of frames drawn from $\mathcal{F}$. In this paper, we consider two cases: 1) the two frames are from the same video clip (intra-video), and 2) the two frames are from different video clips (inter-video). For Case 1), the CIP in a same video clip shows two typical properties: (i) its relative location in the frame does not change much over time, because the camera wearer usually moves his head/eyes to follow the CIP even if the CIP is moving; (ii) The size of the CIP does not change much between neighboring frames. For Case 2), we only consider the synchronized frame pairs from different video clips. In this case, the detected CIP should show consistent 3D motions.

\begin{figure*}[htbp]
\centering \includegraphics[width=1\textwidth]{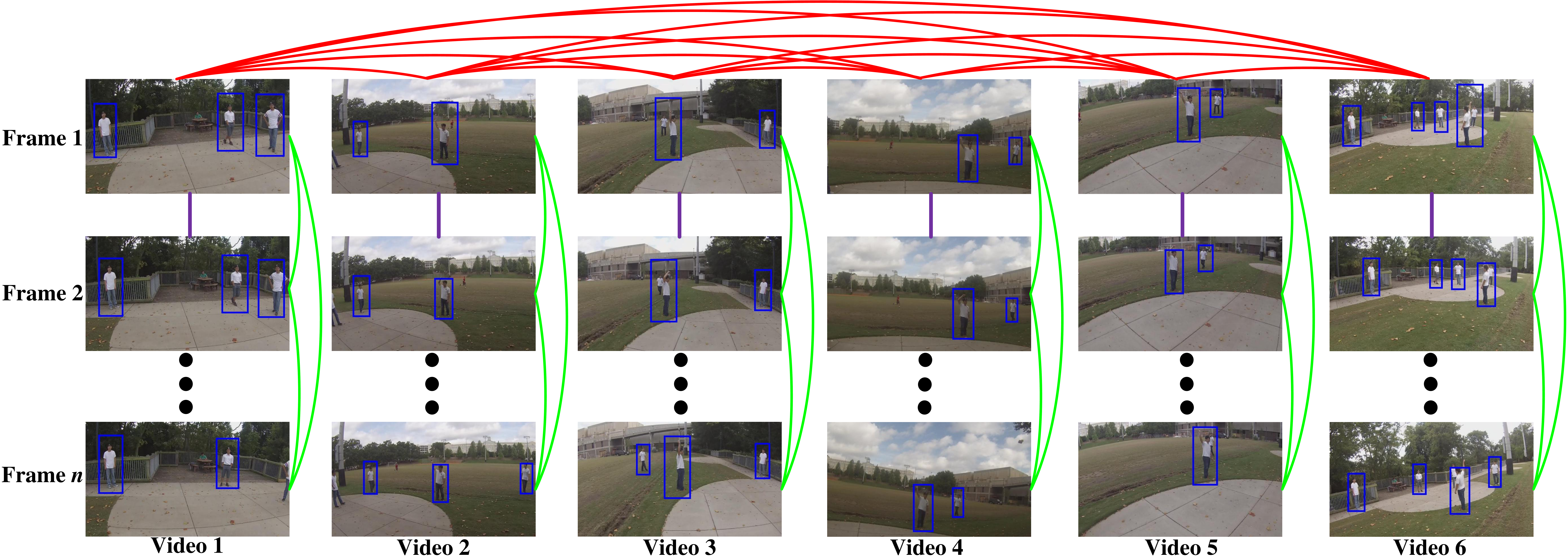} \caption{An illustration of the CRF construction for CIP detection. Each column denotes one video and each row denotes the same frame from different videos. We treat each frame as a node and the detected CIP candidates on each frame as the states of the node. In this CRF, the red lines indicate that the inter-video energies are defined over all pairs of frames between different videos. The green lines indicate that the location-change penalty term in the inter-video energy is defined between each pair of frames inside a video, and the purple lines indicate that the size penalty in the inter-video energy is defined only between neighboring frames inside a video.}

\label{fig:Graph}
\end{figure*}

In our CRF model, we define two different energies $\Psi_{1}$ and $\Psi_{2}$ for the intra-video and inter-video frame pairs, respectively, as illustrated in Fig.~\ref{fig:Graph} and rewrite the energy function $E(H)$ in Eq.~(\ref{eq:CRF}) as
\begin{equation}
E(H)=\sum_{n,t,r\neq t}\Psi_{1}(h_{t}^{n},h_{r}^{n})+\sum_{t,n,m\neq n}\Psi_{2}(h_{t}^{n},h_{t}^{m}).
\label{eq:energyFunction}
\end{equation}
Different from many previous works~\cite{Deselaers12, Fu14}, no unary energy term is defined in this paper since we do not consider the candidate's appearance information. The construction of $\Psi_{1}$ and $\Psi_{2}$ will be elaborated in the following section.




\subsection{Intra-Video Energy and Inter-Video Energy}
\label{sec:energies}
\textbf{\textit{Intra-Video Energy.}}  Ideally, a CIP that draws a camera-wearer's attention usually stays in the view center of the wearer. However, the view center of the wearer may not be perfectly aligned with the center of the camera he/she wears. Therefore, we do not consider center bias in defining the intra-video energy in this work. Instead, the relative location of the CIP usually does not change much in a short video clip and we can penalize the location change between frames for CIP detection. In addition,  in a short video clip, the size of CIP should not change substantially. Considering these two properties, we define the intra-vidoe energy as
\begin{equation}
\begin{split}\Psi_{1}(h_{t}^{n},h_{r}^{n})=1-(\|c_{t}^{n}-c_{r}^{n}\|+1)^{-1}\\
+\delta(t,r-1)\left(1-(\|s_{t}^{n}-s_{r}^{n}\|+1)^{-1}\right)\label{eq:intraVideo}
\end{split}
\end{equation}
where $c_{t}^{n}$, $c_{r}^{n}$ ($s_{t}^{n}$, $s_{r}^{n}$) denote the center (size) of the candidate in frame $t$ and $r$ in video $n$, respectively. $\delta(x,y)$ is the indicator function that equals to 1 if $x=y$ and 0 otherwise. The inclusion of this indicator function ensures that the penalty to the CIP size change is only defined for adjacent frames.

\textbf{\textit{Inter-Video Energy.}} As mentioned above, the inter-video energy is based on motion patterns of the CIP. In this paper, we extract the motion patterns using two types of features: frame-based and trajectory based.

\begin{figure}[htbp]
\centering \includegraphics[width=0.9\columnwidth]{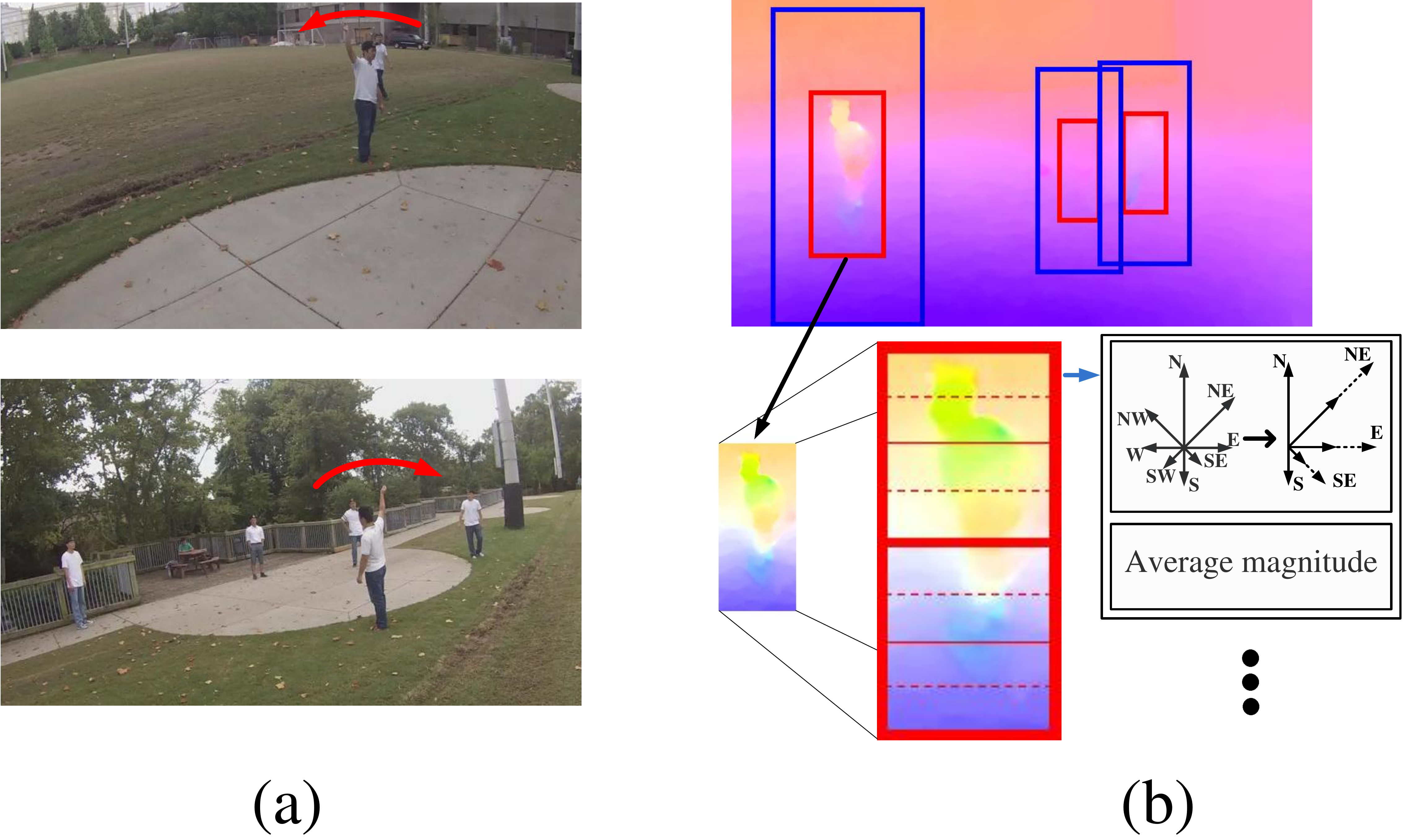} \caption{Frame based motion feature extraction.}
\label{fig:OF}
\end{figure}

The frame-based features are defined to measure the momentary motion of the CIP using the information from
a pair of neighboring frames. Specifically, we calculate the optical flow using neighboring frames~\cite{Brox04}. To remove the influence of camera motion, we further calculate the relative optical flow for each candidate by subtracting the average optical flow in its surrounding region. An example is shown in the top row of Fig.~\ref{fig:OF}(b), where the red box indicates a candidate and the region between the red box and its surrounding blue box is taken for computing the average optical flow for subtraction.

In this paper, we assume all the videos are taken from a similar altitude. This way, at a specific time the a 3D vertical motion of the CIP should be projected to similar directions (up or down) in all the cameras but a 3D horizontal motion may be projected to opposite directions in different cameras. For example, in Fig.~\ref{fig:OF}(a), the same hand motion is from right to left when viewed from front, but from left to right when viewed from back. Therefore, in this paper we propose to ignore the horizontal motion direction information in constructing the frame-based features. Many previous works use a histogram of optical flow (HOF) quantized at 8 directions: East(E), West(W), North(N), South(S), North-East(NE), North-West(NW), South-East(SE) and South-West(SW) as motion features. By ignoring the horizontal motion directions, in this paper, we reduce these 8 directions into 5 by merging three histogram-bin pairs, i.e., merging NW into NE, W into E, and SW into SE, which are vertically symmetric, as shown in Fig.~\ref{fig:OF}(b).

To construct the frame-based features for each CIP candidate on each frame, we divide its bounding box along the vertical direction in a pyramid style, as shown Fig.~\ref{fig:OF}(b). The bounding box is first uniformly divide into two smaller boxes, each of which is then further divided into two equal-size boxes. In our experiment, we perform 3 rounds of pyramid division and in total achieve $1+2+4+8=15$ boxes in 4 scales for each candidate.
By computing and concatenating the 5-bin HOF (as mentioned above) for the original bounding box and the subdivided boxes, we construct an HOF based feature $\hat{f}_{r}^{n}$ with a dimension of $5\times 15=75$.
Within each box (including the original bounding box and its subdivided boxes), we further compute the average magnitudes of the optical flow along $x$ and $y$ directions, and the corresponding standard deviations of these magnitudes along $x$ and $y$ directions, respectively to construct a magnitude based feature $\tilde{f}$ with a dimension of $4\times 15=60$. In this paper, the frame-based feature is defined as the union of the HOF-based and the magnitude-based features.

In practice, the change of the camera angle usually results in the change of the optical-flow magnitudes in $\tilde{f}$. Therefore, when comparing frame-based features between two candidates, we use L1 distance for the HOF-based features and the correlation metric for the magnitude features:
\begin{equation}
\Psi_{F}(h_{t}^{n},h_{t}^{m})=1-\exp(-\|\hat{f}_{t}^{n}-\hat{f}_{t}^{m}\|)+\mathrm{corr}(\tilde{f}_{t}^{n},\tilde{f}_{t}^{m})\label{eq:interVideo_F}.
\end{equation}

In addition to the frame-based features, we also extract trajectory-based features based on short tracklets to capture the motion over a longer time. In this paper, we use Hankelets features and Movement Pattern Histograms (MPH) features for this purpose since both of them show good view-invariance property and have been successfully used for cross-view action recognition~\cite{Li12, Ciptadi14}.

\textbf{Tracklet.} Starting from each candidate, we generate a tracklet with the typical length of 15 frames. In this paper, we use a simple greedy tracking strategy \cite{Zhang14}: given a candidate in a frame, the candidate in the next frame with the highest spatial overlap is taken and this process is then repeated frame by frame to form the tracklet.

\textbf{Dense trajectory.} Improved dense trajectories have been used to efficiently represent videos with camera motions~\cite{Wang13}. In this paper, we extract such improved trajectory features (typically 15 frames). If the majority part of a trajectory, e.g., on more than 8 out of 15 frames, is not coincident with a tracklet, we treat it to be a trajectory in the background. In this paper, we remove background trajectories and only keep the trajectories in the foreground.

\textbf{Hankelet.} Following~\cite{Li12}, we construct one Hankelet (a 16$\times$8 Hankel matrix) for each trajectory. The Hankelets feature for a candidate is the combination of the Hanklets for all the trajectories in this candidate's bounding box.

\textbf{MPH.} The MPH features for a candidate's trajectories consist of 5 histograms, corresponding to the 5 motion directions as used in the frame-based features (see Fig.~\ref{fig:OF}(b)). For each direction, the histogram takes each frame as a bin and the histogram value corresponds to the total trajectory magnitude along this motion direction in this frame.

The difference between two Hankelets $K_r$ and $K_s$ is defined as $d(K_r,K_s)=2-\|K_r K_r^T + K_s K_s^T\|_F$ \cite{Li12}. As mentioned above, each candidate corresponds to a set of Hanklets, one for each trajectory. In this paper, we define the Hankelet based difference between two candidates as the average one over all Hankelet pairs across these two candidates. By using L1 distance for the MPH features, we define the trajectory-based energy term as
\begin{equation}
\small
\Psi_{T}(h_{t}^{n},h_{t}^{m})= \frac{1}{N_H}\sum_{r\in h_{t}^{n};s\in h_{t}^{m}} d(K_r,K_s) + \frac{1}{5}\sum_{d=1}^5 \|M^d_{h_{t}^{n}}-M^d_{h_{t}^{m}}\|
\label{eq:interVideo_T}
\end{equation}
where $N_H$ denotes the number of all different Hankelet pairs across two candidates and $M^d_{h_{t}^{n}}$ indicates the $d$-th histogram (in total 5 directions) in the MPH features.

Finally, we define the inter-video energy as $\Psi_{2}(h_{t}^{n},h_{t}^{m}) = \Psi_{F}(h_{t}^{n},h_{t}^{m}) + \Psi_{T}(h_{t}^{n},h_{t}^{m})$.

\subsection{Identifying the frames without CIP}

\begin{figure}[htbp]
\centering \includegraphics[width=0.8\columnwidth]{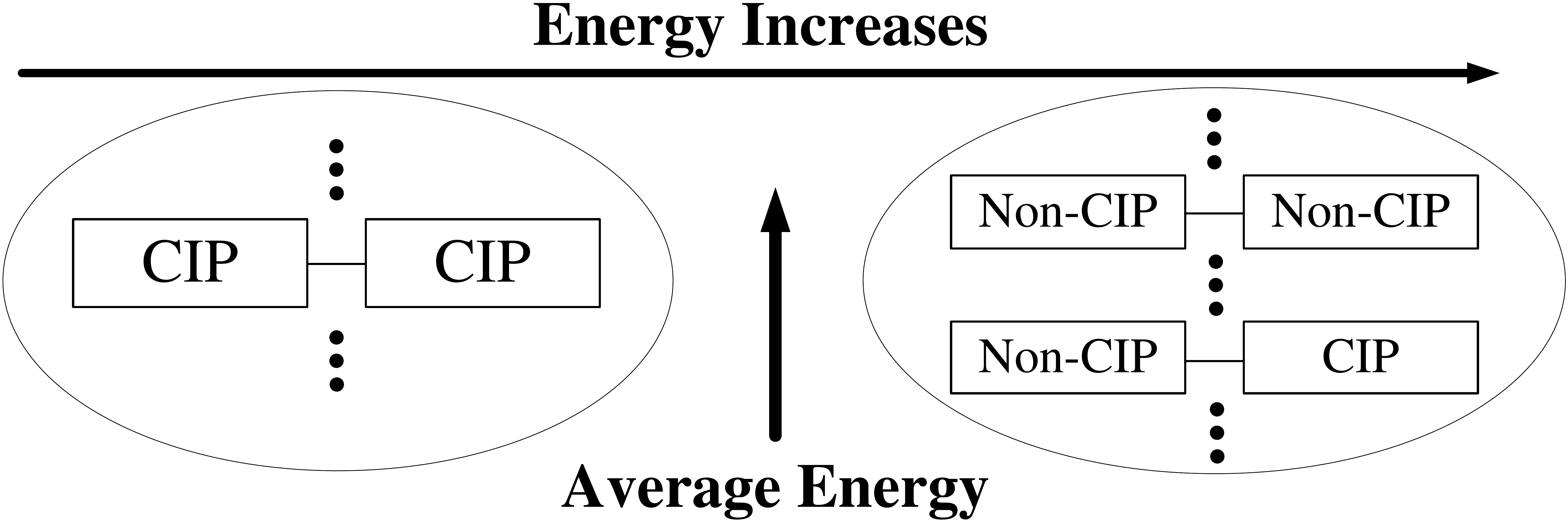}
\caption{An illustration of using average energy over all candidate pairs as the energy terms for idle states. The average energy is located between the minimal energy for a pair of CIPs and the energies between a pair of candidates with at least one non-CIP.
}\label{fig:ego}
\end{figure}

One problem of the CRF model defined above is its assumption that there is always a CIP in each frame. This may not be true in practice. For example, the CIP's egocentric video usually cannot capture himself. Similar issues may occur when the CIP is occluded in some of the frames. To handle this issue, we add an idle state for each node (frame). Let $\mathcal{A}=\{A_{t}^{n}|A_{t}^{n}=h_{t}^{n}\bigcup I_{t}^{n},n=1,\cdots,N;t=1,\cdots,T\}$ denote the state set which includes the idle states $I_{t}^{n}$. The energy function is redefined as
\begin{equation}
E(\mathcal{A})= E(H)+\sum_{n,t,r\neq t}\Psi'_{1}(I_{t}^{n},A_{r}^{n})
+ \sum_{t,n,m\neq n}\Psi'_{2}(I_{t}^{n},A_{t}^{m}),
\label{eq:idleFunction}
\end{equation}
where $\Psi'_{1}$ and $\Psi'_{2}$ denote the intra-video and inter-video energies that involve idle states, respectively. In this paper, we simply define them using the average intra-video energy and inter-video energy over the candidate pairs:
 \begin{equation}
\begin{split} & \Psi'_{1}(I_{t}^{n},A_{r}^{n}|F_{t}^{n},F_{r}^{n})=\frac{1}{W_{1}}\sum_{n,t,r\neq t}\Psi_{1}(h_{t}^{n},h_{r}^{n}|F_{t}^{n},F_{r}^{n})\\
 & \Psi'_{2}(I_{t}^{n},A_{t}^{m}|F_{t}^{n},F_{t}^{m})=\frac{1}{W_{2}}\sum_{t,n,m\neq n}\Psi_{2}(h_{t}^{n},h_{t}^{m}|F_{t}^{n},F_{t}^{m})\label{eq:idleEnergy}
\end{split}
\end{equation}
where $W_{1}$ and $W_{2}$ denote the number of all different candidate pairs used in calculating the average intra-video and inter-video energies, respectively. As illustrated in Fig.~\ref{fig:ego},
the average energy is located between the minimal energy for a pair of CIPs and the energies between a pair of candidates with at least one non-CIP. This will facilitate the selection of idle state in a frame when
the CIP is missing in this frame.

Eq.~(\ref{eq:idleFunction}) is also known as the discrete energy minimization~\cite{Guillaumin13,Kolmogorov06}. In this paper, we use the TRW-S algorithm~\cite{Kolmogorov06} to solve for an approximately optimal solution.

\section{Experimental Results}

\subsection{Data collection}

We collect three sets of temporally synchronized videos taken by multiple wearable cameras. These three sets of videos, denoted as V1, V2 and V3 respectively, are taken in different scenes, including both indoor and outdoor settings. For each video set, there are 6 persons who are both performers and camera wearers and therefore generate 6 videos. Each person wears a GoPro camera over the head. We arrange the video recording in a way that the 6 performers alternately play as the CIP in the video recording by performing different actions. All 6 persons wear white shirts and bluish jeans thus sharing very similar appearances. We manually label the CIP by a bounding box in each frame by using the video annotation tool provided in~\cite{Vondrick12}. In total, we collected 24,000 frames (16 minutes), 25,000 frames (16 minutes 40 seconds) and 20,000 frames (13 minutes 20 seconds) for these three video sets V1, V2, and V3 respectively.


\subsection{Results}
\begin{figure}[htbp]
\centering \includegraphics[width=1\columnwidth]{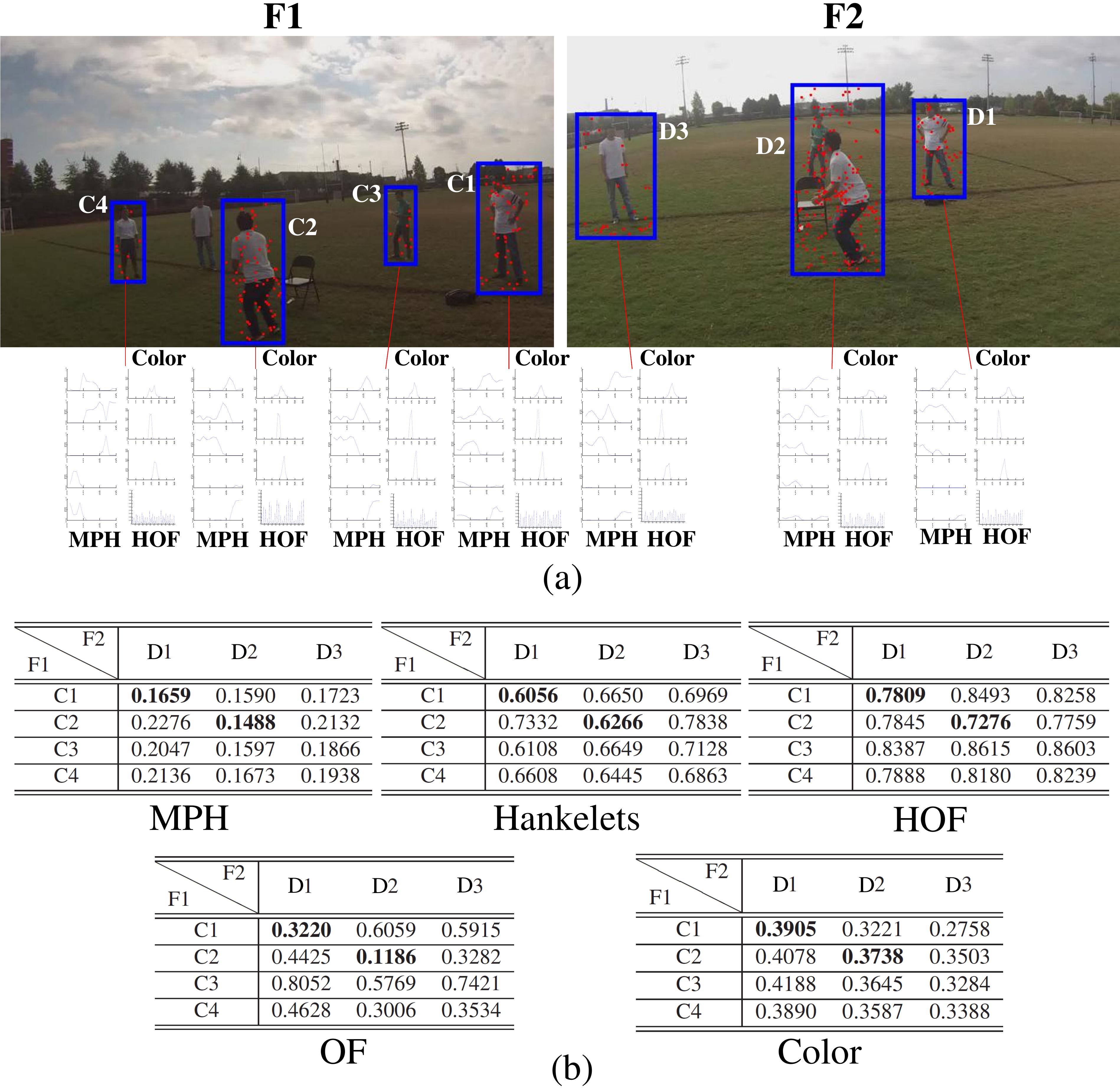}
\caption{An example to illustrate the effectiveness of the proposed motion features.}
\label{fig:simpleCase}
\end{figure}

We first show an example to illustrate the effectiveness of the proposed motion features for identifying the same person from different videos that are temporally synchronized. As shown in Fig.~\ref{fig:simpleCase}(a), blue bounding boxes indicate the detected CIP candidates and red points indicate the improved dense trajectories for each candidate. The MPH features, the color histograms in Lab color channels, and the HOF features are visualized below the corresponding frames. In Fig.~\ref{fig:simpleCase}(b), confusion matrices between different candidates are given when using different features -- each element in the confusion matrices indicates the energy in matching one candidate from frame F1 and a candidate from frame F2. Note that C1 and D1 are the same person, and C2 and D2 are also the same person. Bold font in these matrices indicates the matching energy (i.e., feature difference) of the same person across these two frames and clearly the smaller, the better. We can see that when using the four motion features, these bold-font elements are usually the smallest elements in the respective confusion matrices. However, when using the color features, the bold-font elements are not the smallest in their respective confusion matrix. This shows that the motion features can be more effective than the color features in person identification when the involved people share a very similar appearance.

We then evaluate the proposed algorithm on the collected three video sets. For each detected CIP, denoted by $C$, if there is a ground truth box $G$ with an overlap $O=\frac{C\bigcap G}{C\bigcup G}$ larger than 0.5, we count this detected CIP $C$ to be a true positive. In this way we can calculate the $precision$, $recall$, and the $F$-score$=\frac{2\times precision\times recall}{precision+recall}$.
Table~\ref{tab:performance} shows the quantitative performance of the proposed algorithm and a state-of-the-art video co-segmentation method~\cite{Zhang14}, as well as the variants of the proposed algorithm using different features. For the comparison method~\cite{Zhang14}, instead of using the object proposal result, we directly feed the bounding boxes of the detected candidates to its pipeline. ``Frame based'' and ``Trajectory based'' are the variants of the proposed methods using only the frame-based features and the trajectory-based features, respectively. ``Color based" is another variant of the proposed method using only the color features of Lab histograms instead of any motion features. We can see that the comparison method~\cite{Zhang14} shows a similar performance as ``Color based" and both of them do not perform as good as the proposed algorithm. To demonstrate the usefulness of the location-change penalty term in Eq.~(\ref{eq:intraVideo}), we also report the results of the proposed algorithm without this location-change penalty term, indicated by ``w/o location penalty" in Table~\ref{tab:performance}.

\begin{table}[htbp]
\caption{The performance of the proposed algorithm and its variants, and a comparison video co-segmentation method~\cite{Zhang14}.}
\begin{centering}
\global\long\def\arraystretch{1.3}
 \global\long\def\temptablewidth{0.4\textwidth}
 {\rule{0pt}{1.5pt}}
\small
\begin{tabular}{@{\extracolsep{\fill}}c||c|c|c|c}
\hline
\hline
Methods & Videos  & Precision  & Recall  & $F$-score \tabularnewline
\hline
\multirow{3}*{Method in \cite{Zhang14}} & V1  & 0.4538 & 0.4082 & 0.4298 \tabularnewline
& V2  & 0.4673  & 0.4066  & 0.4348 \tabularnewline
& V3  & 0.4245  & 0.4033  & 0.4136 \tabularnewline
\hline
\multirow{3}*{Color based} & V1  & 0.4232  & 0.4405  & 0.4317 \tabularnewline
& V2  & 0.4401  & 0.4259  & 0.4329 \tabularnewline
& V3  & 0.3812  & 0.4270  & 0.4028 \tabularnewline
\hline
\multirow{3}*{Frame based} & V1  & 0.4667  & 0.5011  & 0.4833 \tabularnewline
& V2  & 0.4481  & 0.5066  & 0.4756 \tabularnewline
& V3  & 0.4089  & 0.4401  & 0.4239 \tabularnewline
\hline
\multirow{3}*{Trajectory based} & V1  & 0.5101  & 0.5523  & 0.5304 \tabularnewline
& V2  & 0.4898  & 0.5396  & 0.5135 \tabularnewline
& V3  & 0.4611  & 0.5122  & 0.4853 \tabularnewline
\hline
\multirow{3}*{w/o location penalty} & V1  & 0.4891  & 0.5207  & 0.5044 \tabularnewline
& V2  & 0.4622  & 0.4758  & 0.4689 \tabularnewline
& V3  & 0.4532  & 0.5107  & 0.4802 \tabularnewline
\hline
\multirow{3}*{Proposed} & V1  & \bf{0.5598}  & \bf{0.6036}  & \bf{0.5809} \tabularnewline
& V2  & \bf{0.5287}  & \bf{0.5682}  & \bf{0.5477} \tabularnewline
& V3  & \bf{0.5027}  & \bf{0.5984}  & \bf{0.5464} \tabularnewline
\hline
\hline
\end{tabular}{\rule{0pt}{1.5pt}}
\par\end{centering}
\label{tab:performance}
\end{table}

Note that the performance of the proposed algorithm is highly dependent on the accuracy of human detection that is used for candidate detection.
If a CIP is present but not detected as a candidate, the proposed algorithm will surely fail to detect the CIP. We also conduct an experiment to evaluate the proposed CIP detection algorithm only on the frames
where the underlying CIP is among the detected candidates. We hope this result can show the performance of the proposed CIP detection by excluding the errors from human detection.
Specifically, if no detected candidate shows a larger-than-0.5 overlap (intersection divided by union) with the ground-truth CIP on a frame, we exclude the CIP detection on this frame from the performance evaluation. Table~\ref{tab:humanDection} shows the results before and after excluding such frames into evaluation.

\begin{table}[htbp]
\caption{The performances of the proposed method before and after excluding the frames where the CIP is present but not among the detected candidates.}
\begin{centering}
\global\long\def\arraystretch{1.3}
 \global\long\def\temptablewidth{0.4\textwidth}
 {\rule{0pt}{1.5pt}}
\small
\begin{tabular}{@{\extracolsep{\fill}}c||c|c|c|c}
\hline
\hline
Models & sets  & Precision  & Recall  & $F$-score \tabularnewline
\hline
\multirow{3}*{Before} & V1  & 0.5598  & 0.6036  & 0.5809 \tabularnewline
& V2  & 0.5287  & 0.5682  & 0.5477 \tabularnewline
& V3  & 0.5027  & 0.5984  & 0.5464 \tabularnewline
\hline
\multirow{3}*{After} & V1  & 0.6134  & 0.6591  & 0.6354 \tabularnewline
& V2  & 0.5960  & 0.6011  & 0.5985 \tabularnewline
& V3  & 0.5789  & 0.6603  & 0.6169 \tabularnewline
\hline
\hline
\end{tabular}{\rule{0pt}{1.5pt}}
\par\end{centering}
\label{tab:humanDection}
\end{table}

\begin{figure*}[htbp]
\centering \includegraphics[width=1\textwidth]{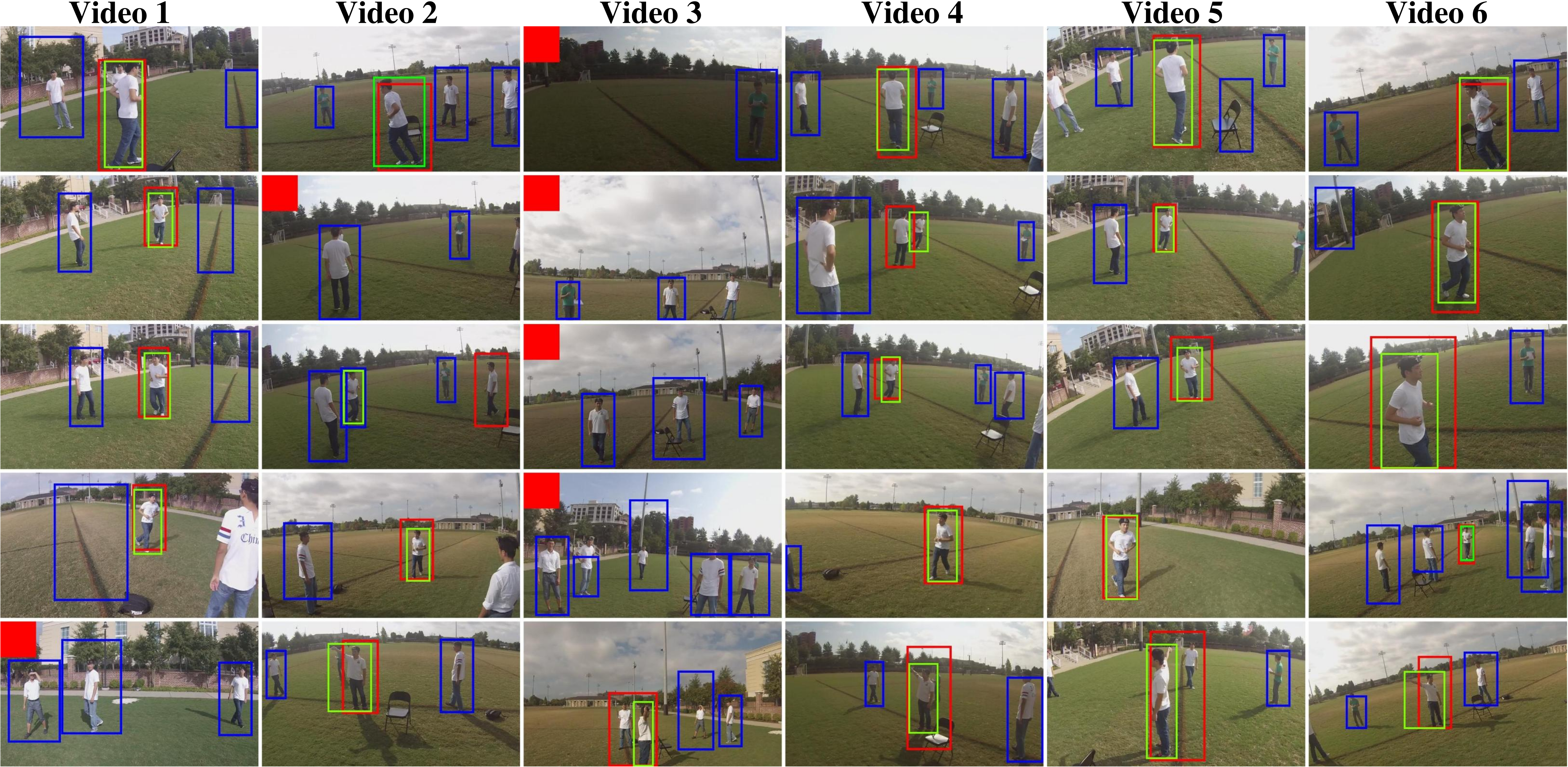}
\caption{The CIP detection on sample frames from V1. 
 Blue, red and green boxes indicate the detected candidates, the detected CIP and the ground truth, respectively. Frames with a solid red square on the top-left corner indicate that no CIP is detected by our algorithm, e.g., they are drawn from the CIP's egocentric video or the CIP is occluded in these frames. Frames with a solid blue square on the top-left corner indicate that no candidate is detected in these frames. Best viewed in color.}
\label{fig:s05}
\end{figure*}

\begin{figure*}[htbp]
\centering \includegraphics[width=1\textwidth]{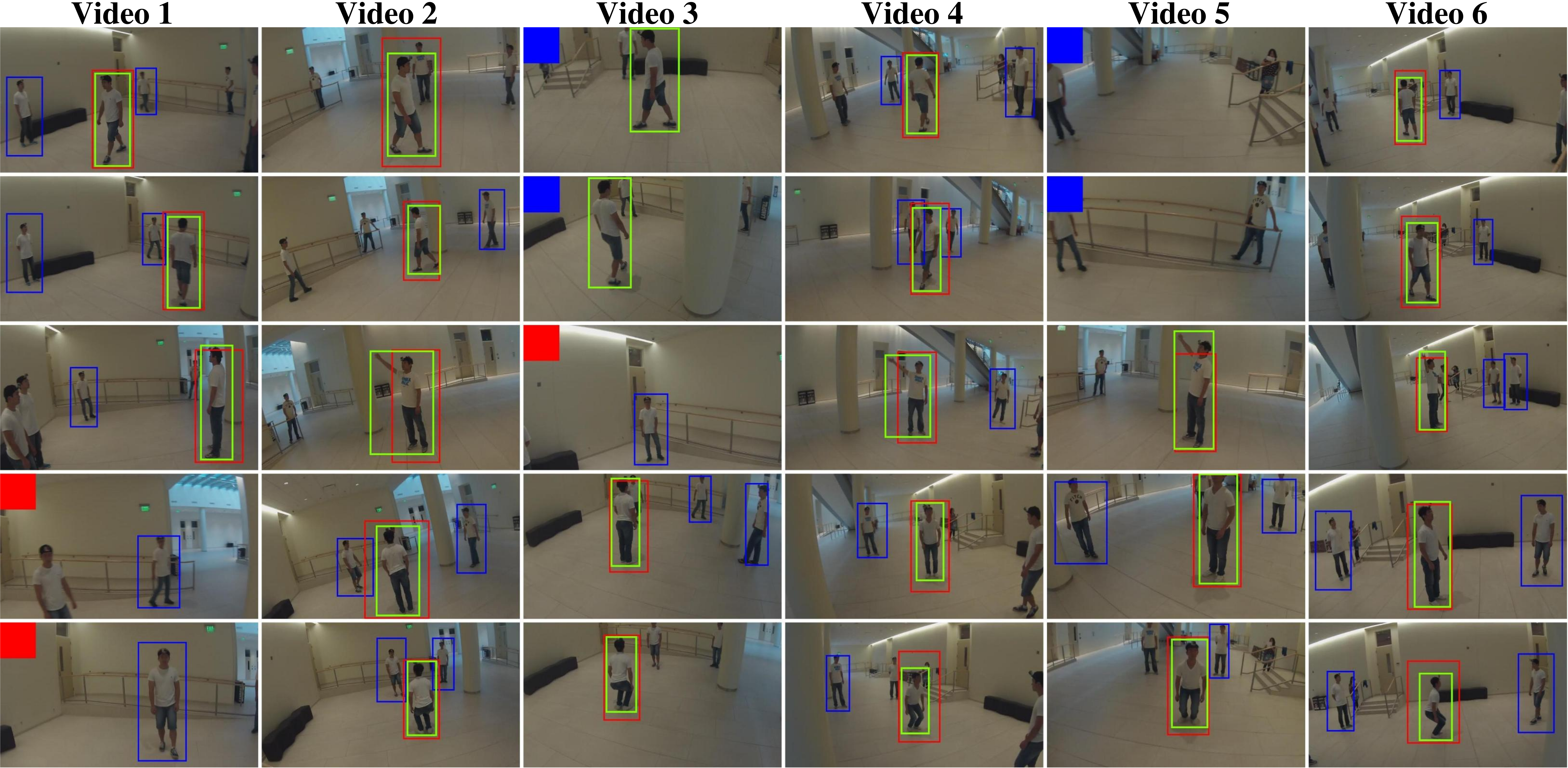}
\caption{The CIP detection on sample frames from V3. See the caption of Fig.~\ref{fig:s05} for the meaning of different-color boxes. Best viewed in color.}
\label{fig:s07}
\end{figure*}

Figures~\ref{fig:s05} and~\ref{fig:s07} show the CIP detection results on sample frames from V1 and V3, respectively. Blue, red and green boxes indicate the detected candidates, the detected CIP and the ground truth, respectively. Frames with a solid red square on the top-left corner indicate that no CIP is detected by our algorithm, e.g., they are drawn from the CIP's egocentric video or the CIP is occluded in these frames. Frames with a solid blue square on the top-left corner indicate that no candidate is detected in these frames. As shown in Fig.~\ref{fig:s05}, the proposed algorithm can detect CIP even if the CIP shows similar appearance to other people in the same scene. From the top four rows of Video 3, the bottom row of Video 1, and the second row of Video 2 in Fig.~\ref{fig:s05}, we can see that the proposed algorithm can handle CIP missing cases, e.g., on the frames drawn from the CIP's egocentric video, by introducing the idle states.
The second row of Video 4 in Fig.~\ref{fig:s05} shows a failure case, which is caused by the partial occlusion of the CIP. The top two rows of Video 3 in Fig.~\ref{fig:s07} show another failure case where the CIP is not detected because it is not among the detected candidates.


The most time consuming steps in the proposed algorithm are the extraction of the raw features, such as the dense trajectories and optical flow. The candidate detection is also time consuming. The major components of the algorithm, including the motion-feature generation, the CRF construction and the CRF optimization, take an average time of 20 seconds (dependent on the number of candidates detected in a video clip) on a laptop with Intel i7-2620M CPU and 4GB RAM, where each CRF is constructed for a 100-frame window over 6 synchronized videos. Therefore, in total 600 frames are modeled by a CRF in our experiments.

\section{Conclusions}

In this paper, we developed a new algorithm to detect co-interest persons (CIPs) from multiple, temporally synchronized videos that are taken by multiple wearable cameras from different view angles. In particular, the proposed algorithm extracts and matches the motion patterns across these videos for CIP detection and can handle the case where the CIP shares a very similar appearance to other nearby non-CIP persons. The proposed algorithm is based on a CRF model which integrates both intra-video and inter-video properties. In the experiments, we collected three video sets, each of which contains six 13+ minute GoPro videos that are temporally synchronized for performance evaluation. The results show that the proposed alglorithm outperforms a state-of-the-art video co-segmentation method and other color-based methods.

\bibliographystyle{abbrv}
{\small
\bibliographystyle{ieee}
\bibliography{activity}
}

\end{document}